\documentclass[letterpaper, 10 pt, journal, twoside]{IEEEtran}

\usepackage{amsmath}
\usepackage{amssymb}
\usepackage{mathtools}
\usepackage{graphicx}
\usepackage{caption}
\usepackage{booktabs}
\usepackage{multirow}
\usepackage{makecell}
\usepackage[normalem]{ulem}

\begin{document}

\title{SACHA: Soft Actor-Critic with Heuristic-Based Attention for Partially Observable Multi-Agent \\Path Finding}

\author{Qiushi Lin and Hang Ma%
\thanks{Manuscript received: February 9, 2023; Revised May 11, 2023; Accepted June 14, 2023.}%
\thanks{This paper was recommended for publication by Editor M. Ani Hsieh upon evaluation of the Associate Editor and Reviewers' comments.
This work was supported by the NSERC under grant number RGPIN2020-06540 and a CFI JELF award.}%
\thanks{The authors are with the School of Computing Science, Simon Fraser University, Burnaby, BC, Canada {\tt\footnotesize \{qiushi\_lin, hangma\}@sfu.ca}}%
\thanks{Digital Object Identifier (DOI): see top of this page.}}

\markboth{IEEE Robotics and Automation Letters. Preprint Version. Accepted June, 2023}%
{Lin \MakeLowercase{\textit{et al.}}: SACHA: Soft Actor-Critic with Heuristic-Based Attention for Partially Observable Multi-Agent Path Finding} 

\maketitle

\begin{abstract}
Multi-Agent Path Finding (MAPF) is a crucial component for many large-scale robotic systems, where agents must plan their collision-free paths to their given goal positions. Recently, multi-agent reinforcement learning has been introduced to solve the partially observable variant of MAPF by learning a decentralized single-agent policy in a centralized fashion based on each agent's partial observation. However, existing learning-based methods are ineffective in achieving complex multi-agent cooperation, especially in congested environments, due to the non-stationarity of this setting. To tackle this challenge, we propose a multi-agent actor-critic method called Soft Actor-Critic with Heuristic-Based Attention (SACHA), which employs novel heuristic-based attention mechanisms for both the actors and critics to encourage cooperation among agents. SACHA learns a neural network for each agent to selectively pay attention to the shortest path heuristic guidance from multiple agents within its field of view, thereby allowing for more scalable learning of cooperation. SACHA also extends the existing multi-agent actor-critic framework by introducing a novel critic centered on each agent to approximate $Q$-values. Compared to existing methods that use a fully observable critic, our agent-centered multi-agent actor-critic method results in more impartial credit assignment and better generalizability of the learned policy to MAPF instances with varying numbers of agents and types of environments. We also implement SACHA(C), which embeds a communication module in the agent's policy network to enable information exchange among agents. We evaluate both SACHA and SACHA(C) on a variety of MAPF instances and demonstrate decent improvements over several state-of-the-art learning-based MAPF methods with respect to success rate and solution quality.
\end{abstract}


\section{Introduction}
\IEEEPARstart{R}{ecent} studies have demonstrated the practical success of Multi-Agent Path Finding (MAPF)~\cite{stern2019multi} in various domains, such as warehouse and office robots~\cite{wurman2008coordinating,veloso2015cobots}, autonomous aircraft-towing vehicles~\cite{morris2016planning}, and other multi-robot systems~\cite{gautam2012review}. MAPF aims to plan collision-free paths for a set of agents from their start positions to their goal positions in a shared environment while minimizing the sum of their completion times (i.e., the arrival times at their goal positions).

Although MAPF is NP-hard to solve optimally~\cite{yu2013structure,banfi2017intractability}, the AI community has developed many optimal and bounded-suboptimal MAPF planners for fully observable environments, where a centralized planner has complete information of the environment to plan joint paths for agents. These planners do not apply to agents with limited sensing capabilities and do not scale well to a large number of agents, as the complexity of coordinating the joint paths of agents grows exponentially with the number of agents in the systems. Learning-based methods with centralized training and decentralized execution have been proposed to develop scalable and generalizable learning-based MAPF methods for the partially observable setting. In this setting, each agent receives a partial observation of its surroundings. Learning-based MAPF methods aim to train a decentralized homogeneous policy that each agent will follow based on its local observation during execution. This policy can be distributed to any number of agents in any environment, as the dimension of the single-agent observation space depends only on the FOV size in the partially observable setting. However, the non-stationarity of environments from the perspective of any single agent poses a significant challenge for learning-based MAPF methods. The transitions of the global state are affected by the individual actions of other agents towards their local interests. Moreover, goal-oriented reinforcement learning with single-agent rewards makes the training process unstable and time-consuming, further incentivizing the selfishness of each agent that prioritizes its goal over collaborating with others. This could hinder coordination and teamwork among agents, negatively affecting overall performance.

To address the challenges posed by solving MAPF in the partially observable setting, we propose \textsc{\textbf{S}oft \textbf{A}ctor-\textbf{C}ritic with \textbf{H}euristic-Based \textbf{A}ttention} (SACHA), a novel approach for partially observable MAPF that leverages heuristic guidance through attention mechanisms to learn cooperation. SACHA builds upon the multi-agent actor-critic framework and, along with its communication-based alternative, SACHA(C), aims to learn a decentralized homogeneous policy that can be generalized to any number of agents in any arbitrary partially observable MAPF environment. To achieve this, we first allow each agent to access the goal-oriented heuristic guidance of multiple agents in the form of the shortest path distances to each of the goals, which can be computed efficiently before execution. We then employ a self-attention module in the policy network for each agent to locally select relevant information from the guidance and take actions towards better cooperation among agents.

We expect SACHA to make a significant algorithmic impact not only on MAPF solving but also on other similar multi-agent tasks in partially observable settings because its learning process of the homogeneous policy is also guided by a homogeneous critic for more stable learning and faster convergence. Unlike existing multi-agent actor-critic methods with one fully centralized critic or multiple decentralized critics, SACHA introduces a novel agent-centered critic network that uses an attention mechanism to approximate each agent's $Q$-function and performs credit assignment only based on the information within each partial observation. The input dimension of this critic is determined by the number of agents that each agent's $Q$-function should be based on, which is implicitly limited by the partial observation range (e.g., FOV size in MAPF). This partially centralized critic ensures that the $Q$-function is not biased towards any specific problem instance, resulting in a well-trained policy network that can generalize well to different numbers of agents and environments.

We experimentally compare SACHA and SACHA(C) with state-of-the-art learning-based and search-based MAPF methods over several MAPF benchmarks. Our results show that both versions of SACHA result in higher success rates and better solution quality than other methods in almost all test cases. The results thus indicate that our methods allow for better cooperation among agents than the other methods with and even without communication.

\section{Problem Formulation}
In this section, we first present the standard formulation of MAPF and then dive into its partially observable variant.

\subsection{Standard MAPF Formulation}
In the standard formulation of MAPF, we are given a connected and undirected graph $G=(V, E)$ and a set of $M$ agents, indexed by $i \in \{1, 2, \cdots, M\}$. Each agent has a unique start vertex $s_i \in V$ and a unique goal vertex $g_i \in V$. Time is discretized into time steps, $t = 0, 1, \cdots, \infty$. Between two consecutive time steps, each agent can either \textit{move} to one of its adjacent vertices or \textit{wait} at its current vertex. A path for agent $i$ contains a sequence of vertices that lead agent $i$ from $s_i$ to $g_i$, where each vertex indicates the position of the agent for every time step. The \textit{completion time} $T_i$ of agent $i$ is defined as the length of its path, and it is the earliest time when agent $i$ has reached and terminally stayed at its goal vertex. Collisions between agents are not allowed. A vertex collision occurs when two agents occupy the same vertex $v$ at the same time $t$. An edge collision occurs when two agents traverse the same edge $(u, v)$ in opposite directions from $t$ to $t+1$. A MAPF solution is a set of collision-free paths for all agents. A commonly-used objective function is the average (equivalently, sum) of the completion times of all agents.

\begin{figure}[t]
\centering
\includegraphics[width=0.88\linewidth]{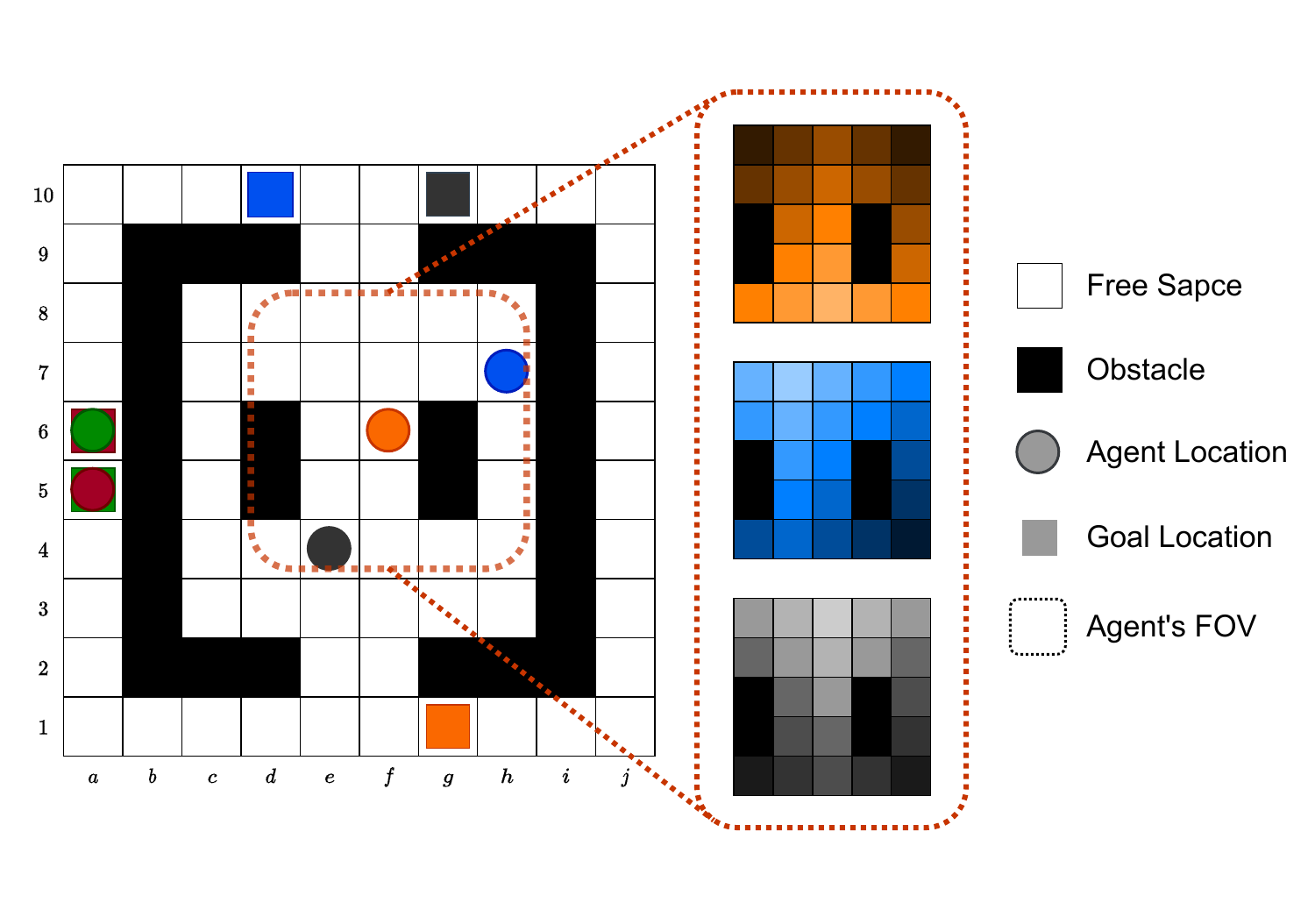}
\caption{
A partially observable MAPF instance and the multi-agent heuristic maps for the orange agent. Each agent's current (circle) and goal (square) positions have the same color. In this instance, each agent's FOV is a $5 \times 5$ area. The orange agent's FOV contains three agents, including itself. Therefore, the policy network of the orange agent can potentially utilize three heuristic maps for the same $5 \times 5$ area. A darker shade of color represents a greater distance to the goal.
}
\label{fig:pomapf}
\end{figure}

\begin{table*}[t]
\centering
\caption{Comparison of learning-based methods for partially observable MAPF}
\label{tab:related_work}
\scalebox{0.88}{%
\begin{tabular}{|l|c|c|c|c|}
\hline
\multicolumn{1}{|c|}{Method}       & Learning Framework               & Communication & Single-Agent Guidance   & Cooperative Guidance                     \\ \hline
PRIMAL \cite{sartoretti2019primal} & A3C (RL) + Behavior Cloning (IL) & Inapplicable  & Goal Direction          & Goal Directions of Neighbouring Agents         \\
DHC \cite{ma2021distributed} & IQL & Required & Shortest Path Distances & $-$ \\
DCC \cite{ma2021learning}    & IQL & Required & Shortest Path Distances & $-$ \\
SACHA (Ours)                       & Multi-Agent Soft Actor-Critic         & Optional      & Shortest Path Distances & Shortest Path Distances of Neighbouring Agents \\ \hline
\end{tabular}%
}%
\end{table*}

\subsection{Partially Observable Environments}
In this paper, we consider a more practical scenario where agents have only partial observation of the environment but still aim to fully cooperate to minimize the average completion time. We model this partially observable variant of MAPF as a decentralized partially observable Markov Decision Process (Dec-POMDP)~\cite{littman1994markov}, defined as a 7-tuple $\langle \mathcal{S}, {\mathcal{A}_i}, P, {\Omega_i}, O, R, \gamma \rangle$. $\mathcal{S}$ is the set of global states. $\mathcal{A}_i$ is the set of actions for agent $i$, and $\mathcal{A} = \prod_{i=1}^{M}\mathcal{A}_i$ is the joint action space. $\Omega_i$ is the observation space of agent $i$, and $\Omega = \prod_{i=1}^{M}\Omega_i$ is the joint observation space. $\mathcal{O}: \mathcal{A} \times \mathcal{S} \rightarrow \Omega$ is the observation function, denoting the probability $P(o|a, s)$, whereas $\mathcal{P}: \mathcal{A} \times \mathcal{S} \rightarrow \mathcal{S}$ is the state-transition function for the environment, representing the probability $P(s' | a, s)$, where $o \in \Omega$, $a \in \mathcal{A}$, and $s, s' \in \mathcal{S}$. $\mathcal{R}: \mathcal{S} \times \mathcal{A} \rightarrow \mathbb{R}$ is the reward function, and $\gamma$ is the discount factor.

In MAPF, the observation and state-transition functions are deterministic, where each agent has full control of its next position and observation by taking one of the move actions or the wait action. To facilitate proper comparison with existing learning-based MAPF methods, we formalize the MAPF problem on two-dimensional grid maps with four neighbors, even though our method can also be generalized to other MAPF problems. The partial observability limits each agent's perception to its FOV, defined as a $\mathcal{L} \times \mathcal{L}$ square area centered on the agent. Agents take their actions based on their local observation and the history from the beginning to the current time step.

One of the key challenges for decentralized planners with limited access to global information is the occurrence of deadlocks and livelocks. Similar to time-windowed MAPF planners~\cite{li2021lifelong}, these issues arise due to the limited planning horizon of agents, either in time or space, that prevents them from reaching their goals. For instance, consider Fig.~\ref{fig:pomapf}, where the red agent in $a5$ and the green agent in $a6$ are heading towards their respective goals $a6$ and $a5$. An optimal solution may require the red agent to move north and terminally stay at its goal while the green agent moves north and takes a detour since its direct path is blocked by the red agent. However, after a few moves, the green agent will no longer observe the blocking red agent and end up wiggling between $a9$ and $a10$ indefinitely. Symmetrically, if the green agent moves south and terminally stays at its goal, the red agent will eventually wiggle between $a1$ and $a2$.

Existing learning-based MAPF methods rely on two extra assumptions to alleviate the above issues: (1) Each agent has full visibility of the map (which is consistent with both standard MAPF and time-windowed MAPF), even though it does not know the global state. (2) Two agents are allowed to communicate when they are within each other's FOV. In this paper, SACHA and SACHA(C) utilize the same assumptions. Both methods give each agent access to the shortest path distances to its goal. SACHA(C) enables inter-agent communication, while SACHA only requires each agent to identify other agents in its FOV.

\section{Related Work}
We now survey the related work on learning-based MAPF methods and multi-agent reinforcement learning methods.

\subsection{Learning-Based MAPF Methods}
Recent learning-based MAPF methods~\cite{sartoretti2019primal,liu2020mapper,ma2021distributed,ma2021learning} have been proposed to solve MAPF in a partially observable setting. These methods aim to learn a decentralized policy that can be generalized to different MAPF instances. While centralized MAPF planners require full observation of the environment and must plan paths from scratch for each instance, the well-trained model can be applied to MAPF instances with any number of agents and environment size, without increasing the time complexity.

The most straightforward approach for tackling partial observability is to treat other agents as part of the environment and let each agent learn its policy independently, as in Independent $Q$-Learning (IQL)~\cite{tan1993multi}. However, this approach results in non-cooperative behavior among the agents, and its training is not guaranteed to converge due to interference between the policies of different agents. State-of-the-art MAPF methods~\cite{ma2021distributed,ma2021learning} enhance IQL with a communication mechanism to promote cooperation between agents. Other MAPF methods~\cite{sartoretti2019primal,liu2020mapper} use the actor-critic framework, with guidance from an expert demonstration. PRIMAL~\cite{sartoretti2019primal} combines on-policy asynchronous advantage actor-critic (A3C)~\cite{mnih2016asynchronous} with behavior cloning from an expert demonstration generated by a centralized MAPF planner~\cite{wagner2011m}. However, the centralized MAPF planner requires solving numerous MAPF instances, which slows down the training process. DHC~\cite{ma2021distributed} and DCC~\cite{ma2021learning} have shown that using single-agent shortest path distances as heuristic guidance for goal-oriented learning of each agent is more effective than following a specific reference path in a multi-agent cooperative setting.

In Section~\ref{sec:experiments}, we compare SACHA against PRIMAL, DHC, and DCC experimentally. Table~\ref{tab:related_work} summarizes the comparison of properties of these methods, showing that SACHA improves over them by adopting a more stable training scheme, utilizing better heuristic guidance through more complex model design, and allowing for applicability to both communicating and non-communicating scenarios.

\subsection{Cooperative Multi-Agent Reinforcement Learning}
Multi-Agent Reinforcement Learning (MARL) is a well-established framework for coordinating multiple agents in a shared environment. A rich literature~\cite{tampuu2017multiagent,gupta2017cooperative,oroojlooy2022review} on cooperative MARL has been dedicated to coordinating agents that work towards a common objective and take actions that benefit all agents as a whole. To deal with the nonstationarity of the environment, most existing actor-critic methods use one or more fully centralized critics that observe the entire environment. For example, Multi-Agent Deep Deterministic Policy Gradient \cite{lowe2017multi} trains each actor with only its local observation using the DDPG algorithm~\cite{lillicrap2015continuous}, while its corresponding fully centralized critic can access the observations and actions of all agents. Instead of using multiple fully centralized critics, Counterfactual Multi-Agent Policy Gradient~\cite{foerster2018counterfactual} uses only one fully centralized critic that learns to assign credit to agents and estimate $Q$-functions for all agents based on a counterfactual baseline that marginalizes out the action of each individual agent. However, it becomes increasingly difficult to perform such credit assignments for cases with large numbers of agents. Therefore, Multiple Actor Attention-Critic\cite{iqbal2019actor} deploys an attention mechanism for the fully centralized critic to selectively pay attention to relevant information from all agents. SACHA also uses a similar attention mechanism but differs from existing actor-critic methods by using a novel homogeneous agent-centered critic that only takes the local information from each agent as input for generalizability to different MAPF instances instead of a fully centralized critic specific to only one MAPF instance.

\section{SACHA}
We now provide a detailed description of the main components of SACHA. First, we describe how we use the shortest path distances of multiple agents as the cooperative guidance for each individual agent. Next, we explain how we integrate the shortest path distances into the commonly applied reward design. We also elaborate on our new model design, which includes attention mechanisms applied to both the actors and the critic and an optional communication module. Lastly, we discuss SACHA in the context of the multi-agent actor-critic learning framework.

\subsection{Multi-Agent Shortest Path Distance Heuristic Maps}

Empirical studies \cite{ma2021learning} have shown that shortest path distances from all vertices to each agent’s goal vertex can greatly benefit goal-oriented learning for the agent in partially observable environments. SACHA utilizes multi-agent shortest path distances to guide not only the achievement of single-agent goals but also better cooperation between agents. Specifically, a backward uniform-cost search is run from each agent’s goal vertex to all vertices in the given graph to generate the shortest path heuristic maps for the agent. The heuristic maps for all agents can be calculated offline once the graph and all goal vertices are given. They remain unchanged for the same MAPF instance during training and can be efficiently generated in advance for a new MAPF instance during execution. The time complexity for calculating the distances for $M$ agents on graph $G=(V, E)$ is $\mathcal{O}(M |E| \log |V|)$. Many search-based and some recent learning-based MAPF methods \cite{sharon2015conflict,ma2021distributed,ma2021learning} also use heuristic maps of shortest path distance, but only as single-agent guidance. SACHA gives each agent access to not only its heuristic map but also the heuristic maps of other agents within its FOV, which enables better cooperation. Fig.~\ref{fig:pomapf} visualizes the heuristic maps that the orange agent has access to at the current time step. Since there are three agents within its FOV, three corresponding heuristic maps are input to the agent's policy network that we will describe later.

\begin{table}[t]
\centering
\caption{Individual reward function from DHC~\cite{ma2021distributed}.}
\label{tab:reward_design}
\begin{tabular}{|c|c|}
\hline
\textbf{Action}                 & \textbf{Reward} \\ \hline
Move (up / down / left / right) & -0.075          \\ \hline
Wait (on goal, away goal)       & 0, -0.075       \\ \hline
Collision (obstacles or agents) & -0.5            \\ \hline
Reaching Goal                   & 3               \\ \hline
\end{tabular}%
\end{table}

\subsection{Reward Design with Heuristic Maps}
We design the reward function for each individual agent based on its heuristic map. We start with the individual reward function of DHC~\cite{ma2021distributed} as shown in Table~\ref{tab:reward_design}. It follows the intuition that an agent is punished slightly for each time step before arriving at the goal, thereby encouraging it to reach its goal as quickly as possible. To improve the success rate of solving the global MAPF task, each agent is punished to a greater extent each time it collides with another agent or an obstacle. The agent receives a positive reward only when it arrives at its goal. Most existing learning-based MAPF methods follow the same design idea for their reward functions. However, the goal-conditioned reward in this reward design makes the training unstable and difficult to converge, especially for long-horizon tasks such as MAPF. Also, since an agent which stays further away from its goal generally has a larger potential to collide with other agents, the move actions for it should not be rewarded the same as for ones that are closer to their goals. Therefore, motivated by Heuristic-Guided Reinforcement Learning (HuRL)~\cite{cheng2021heuristic}, we reshape the reward function for each agent with an additional heuristic term. Assume we have a transition tuple, $(s, a, r, s')$, we reshape the reward as:
\begin{equation}
\tilde{r}_i(s, a) = r_i(s, a) + (1-\lambda) \gamma h_i(s'),
\end{equation}
where $h_i(s')$ is the negated normalized shortest path distance of the global state $s'$ from the heuristic map of agent $i$. This heuristic term represents a priori guess of the desired long-term return of an agent from state $s'$ and thus serves as a horizon-based regularization. HuRL has been proved both theoretically and empirically to be able to accelerate the learning process significantly by intrinsically reshaping the reward of every position for each agent. We set $\lambda$ to 0.1 in the experiments.

\subsection{Model Design with Attention Mechanisms}

\begin{figure*}[t]
\centering
\includegraphics[width=0.88\linewidth]{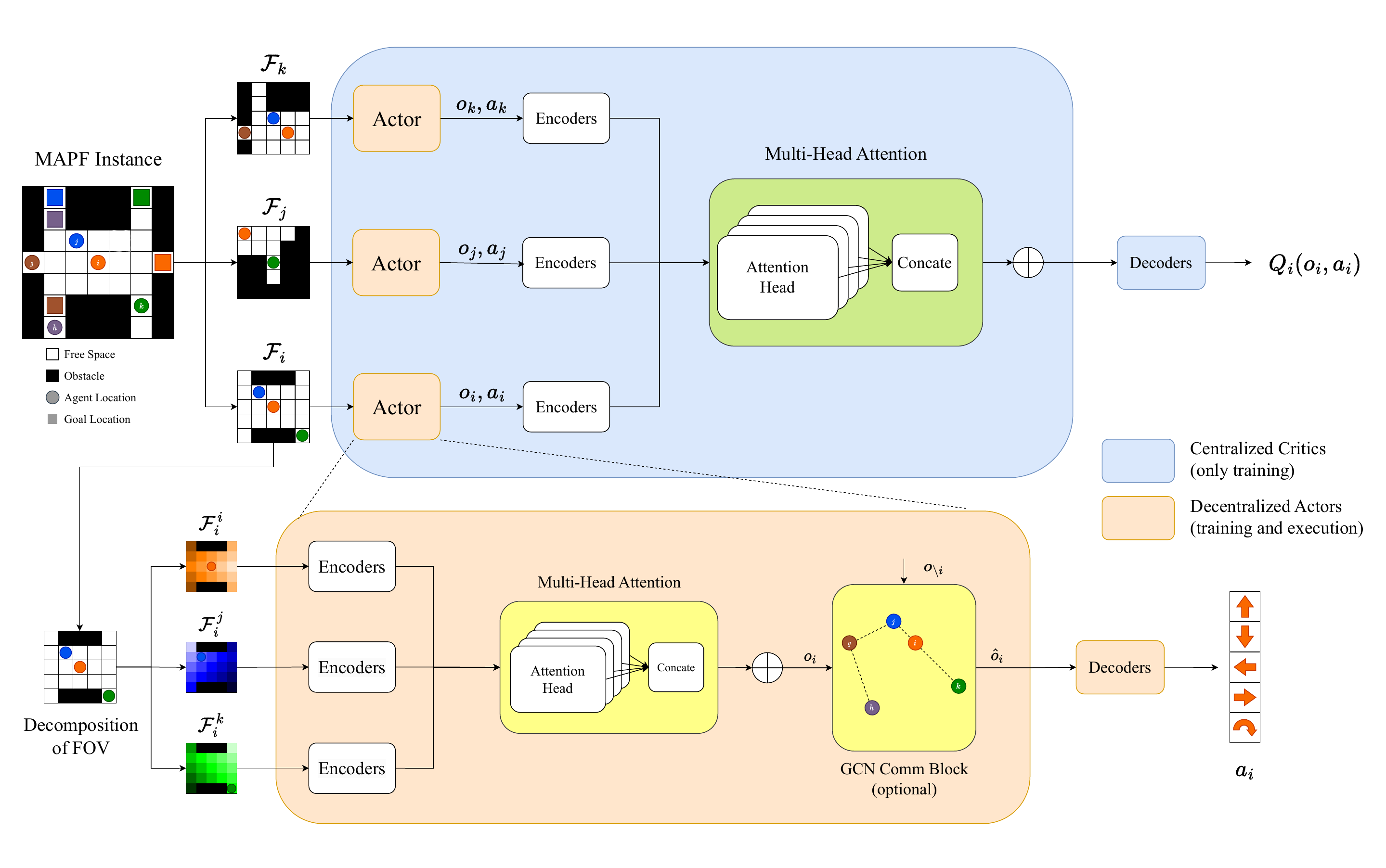}
\caption{
Model design of SACHA and SACHA(C). The framework consists of the actors (red) and the critics (blue). The bottom part demonstrates agent $i$'s policy network. Within its $5 \times 5$ FOV, there are three agents inside and hence three heuristic maps are fed into its network. After going through the linear functions, three output features will then be processed by the attention block, picking out the relevant information required for agent $i$ to take its action. The upper part shows how the critic assigns the credit for agent $i$ based on the local observations and the corresponding actions from agents in its subgroup via a similar attention mechanism.
}
\label{fig:model_design}
\end{figure*}

We propose a novel model architecture based on the multi-agent actor-critic framework. Our model aims to achieve generalization across different instances by restricting the actor and the critic to operate only within the observation of each agent. At each time step $t$, we define an undirected observation graph $G_t = (V, E_t)$, where $V$ is the set of all agents and each edge in $E_t$ indicates that the corresponding agents can observe each other. The time-varying graph $G_t$ captures the dynamic correlation of agents in partially observable environments. We denote the subgroup centered on agent $i$ as $\{i \cup \mathcal{N}_t(i)\}$, where $\mathcal{N}_t(i)$ is the set of the nearest $K - 1$ neighbors of node $i$ inside its FOV. The observation of each agent $i$ consists of a set of $K$ feature maps $\mathcal{F}_i = \{\mathcal{F}_i^j\}_{j \in i \cup \mathcal{N}_t(i)}$ where $\mathcal{F}_i^j \in \mathbb{R}^{\mathcal{L} \times \mathcal{L} \times 3}$. Each feature map in the set corresponds to an agent in the subgroup of agent $i$ and contains three channels: (1) a binary matrix that identifies the obstacles and the free space, (2) a binary matrix that marks the positions of agents, and (3) a heuristic channel that shows the normalized shortest path distances for each empty cell. $K$ is set to $3$ in our experiments.

We present the learning framework of SACHA in Figure~\ref{fig:model_design}. Given the observation features, the policy network starts with the observation encoders with shared parameters. The encoders consist of several convolutional layers followed by a GRU~\cite{cheng2021heuristic} memory unit. The output set of $K$ encoding is input into the Multi-Head Attention (MHA)~\cite{vaswani2017attention} module that learns the interaction between agent $i$ and its subgroup members by selectively attending to relevant information. The MHA module outputs a set of features the sum of which is used for the observation representation, denoted as $o_i$. It then will be passed to decoders to generate the corresponding action vector $a_i \in \mathbb{R}^5$. Each element in $a_i$ represents the probability of choosing one of the five discrete actions $\{up, down, left, right, stay\}$.

We propose a novel agent-centered critic, that evaluates each agent's action individually based on its local observation and information about its subgroup members. Unlike previous methods that use a centralized critic with global information, our critic leverages the attention mechanism to dynamically assign credit to each agent. We first pass the policy network's output through a linear function and then apply a multi-head attention module. The sum of the concatenated output vectors is then forwarded through the decoders to obtain the final $Q$-value, which is used to update the policy networks via the policy gradient method. Since our agent-centered critic takes only requires the local information of the central agent, it is not dependent on any specific environment information, and the learned policy network can thus generalize better.

\subsection{Optional Communication Module}
Furthermore, we propose a communication-based variant of our method, named SACHA(C). To encourage better cooperation, our method should be able to take advantage of communication when it is allowed. We add an optional communication module after the multi-head attention block, as shown in Fig.~\ref{fig:model_design}. We first gather all observation representation $\{o_i\}_{i=1}^{M}$ as the initialization ,$H^{(0)}$, and then feed it into a multi-layer Graph Convolution Network~\cite{kipf2016semi} (GCN). Recalled that $M$ is the number of agents. Let $A_t$ be the adjacency matrix of $G_t$ and $\tilde{A}_t = A_t + I_M$. Define $\tilde{D}_t$ as a diagonal matrix where $\tilde{D}_{ii} = \sum_{j} \tilde{A}_{ij}$. The output of the $(l+1)$-th layer would be:
\begin{equation}
H^{(l+1)} = \sigma(\tilde{D}_t^{-\frac{1}{2}} \tilde{A}_t \tilde{D}_t^{-\frac{1}{2}} H^{(l)} W^{(l)}),
\end{equation}
where $\sigma(\cdot)$ is the sigmoid function and $W$ is a layer-specific trainable weight matrix. After $l$ layers of GCN, we can decompose $H^{(l+1)}$ to $M$ corresponding vectors, $\{\hat{o}_i\}_{i=1}^{M}$, which will eventually be decoded by each network to their corresponding action vectors as usual. We choose a two-layer GCN in the SACHA(C). The communication module is optional, but it can make agents reach information outside their local observation and hence achieve better cooperation.

\subsection{Soft Actor-Critic with Multi-Agent Advantage Function}
SACHA updates the agent's policy network $\pi$ parameterized by $\theta$ and the critic network parameterized by $\psi$ simultaneously through the soft actor-critic framework. We let $\bar{\theta}$ and $\bar{\psi}$ denote the moving average of $\theta$ and $\psi$ (target parameters of the actor and the critic network), respectively. We first define the action-value temporal difference (TD) error for any experience, $e = (s, a, \tilde{r}, s')$, from the replay buffer $D$:
\begin{align}
\delta_i = \, &Q_i^{\psi}(o_i, a_i) - \tilde{r}_i \nonumber \\ 
&- \gamma \mathbb{E}_{a_i' \sim \pi_{\bar{\theta}}(o_i')} [ Q_i^{\bar{\psi}}(o
_i', a_i') - \alpha \log (\pi_{\bar{\theta}}(a_i'|o_i'))) ]
\end{align}
where $\alpha$ is the temperature parameter that decides the weight of the entropy term in the soft actor-critic framework~\cite{haarnoja2018soft}. SACHA runs the critic network through every agent-centered subgroup and updates it by minimizing the mean square error loss function:
\begin{equation}
L_Q(\psi) = \mathbb{E}_{e \sim D} \displaystyle \sum_{i=1}^{M(e)} \frac{\delta_i^2}{M(e)},
\end{equation}
where $M(e)$ is the number of agents in $e$. On the other hand, the actors update their underlying policy networks by the policy gradient via the $Q$-values from the critic network:
\begin{align}
\nabla_{\theta_i} J (\theta) = \mathbb{E}_{o_i \sim D, a_i \sim \pi_{\theta_i}(o_i)} [ \nabla_{\theta_i} \log(\pi_{\theta_i}(a_i | o_i)) \nonumber \\
(Q_i^{\psi}(o_i, a_i) - b(o_i, a_{\setminus i}) -\alpha \log(\pi_{\theta_i}(a_i | o_i))) ]
\label{eq:local_pg}
\end{align}
where $\nabla_{\theta} \log(\pi_{\theta}(a_i | o_i))$ is the score function and $Q_i^{\psi}(o_i, a_i) - b(o_i, a_{\setminus i})$ is the multi-agent advantage function. Inspired by COMA \cite{foerster2018counterfactual}, SACHA adopts the counterfactual baseline in a discrete action space as follows, which marginalizes out the specific action of agent $i$:
\begin{equation}
b(o, a_{\setminus i}) = \displaystyle \sum_{a'_i \in \mathcal{A}_i} \pi(a'_i|o_i) Q_i^{\psi}(o_i, (a'_i, a_{\setminus i})),
\end{equation}
where $a_{\setminus i} \in \mathcal{A}_{\setminus i} = \prod_{j \neq i} \mathcal{A}_j$ is the combination of actions from all agents except for $i$ and $\mathcal{A}_i$ is each agent's action space. This baseline specifically compares an action to other actions of agent $i$ by fixing the actions of all other agents and invoking the critic network for $|\mathcal{A}_i|$ times. In each instance, we collect these $M$ updated policies and aggregate them into the new policy by averaging all the locally updated policies: $\theta^{(t+1)} = \sum_{i=1}^{M} \frac{\theta_i^{(t)}}{M}$, where $\theta_i^{(t)} = \theta^{(t)} - \nabla_{\theta_i} J (\theta^{(t)})$. The policy and the critic network are updated together iteratively to reach fast and stable convergence.

\section{Analysis}
\label{sec:analysis}
In this section, we analyze the effectiveness of the policy gradient in our multi-agent actor-critic framework. Most of the existing learning-based MAPF methods use IQL, in which each agent treats others as part of the environment and updates the global policy $\theta$ as follows:
\begin{equation}
\nabla_{\theta} J (\theta) = \mathbb{E}_{s \sim D, a \sim \pi_{\theta}}[\nabla_{\theta} \log(\pi_{\theta}(a | s)) Q(s, a)],
\end{equation}
where $s$ and $a$ denote the joint state and joint action, respectively. Now we show that this gradient is equivalent to Eq.~\eqref{eq:local_pg}. We omit the entropy term here, but the proof can be easily extended to include it.

Since each agent acts on its policy independently, we have $\pi_{\theta}(a|s) = \prod_i \pi_{\theta_i}(a_i|s)$. We represent the stationary distribution induced by $\pi_{\theta}$ as $d_{\theta}(s)$, meaning the probability of being in the state $s$ by following $\pi_{\theta}$. Following the proof in \cite{sutton1999policy}, we get:
\begin{align*}
& \nabla_{\theta_i} J (\theta) \\
& = \displaystyle \sum_{s \in D} d_\theta(s) \sum_{a \in \mathcal{A}} \pi_\theta(a | s) \nabla_{\theta_i} \Bigl[ \sum_{j=1}^M \log(\pi_{\theta_j}(a_j | s)) \Bigr] Q(s, a) \\
& = \displaystyle \sum_{s \in D} d_\theta(s) \sum_{a \in \mathcal{A}} \pi_\theta(a|s) \nabla_{\theta_i} \log(\pi_{\theta_i}(a_i | s))  Q(s, a) \\
& = \displaystyle \sum_{s \in D} d_\theta(s) \sum_{a \in \mathcal{A}} \Bigl[ \prod_{j \neq i} \pi_{\theta_j}(a_j | s) \Bigr] \nabla_{\theta_i} \pi_{\theta_i}(a_i | s) Q(s, a).
\end{align*}
To further the proof, we consider the following equation:
\begin{align*}
& \displaystyle \sum_{a \in \mathcal{A}} \Bigl[ \prod_{j \neq i} \pi_{\theta_j}(a_j | s) \Bigr] \nabla_{\theta_i} \pi_{\theta_i}(a_i | s) F(s, a_{\setminus i}) \\
= & \displaystyle \sum_{a_{\setminus i} \in \mathcal{A}_{\setminus i}} \Bigl[ \prod_{j \neq i} \pi_{\theta_j}(a_j | s) \Bigr] F(s, a_{\setminus i}) \Bigl[ \nabla_{\theta_i} \underbrace{\displaystyle \sum_{a_i \in \mathcal{A}_i} \pi_{\theta_i}(a_i | s)}_{=1} \Bigr] \\
= & \, 0.
\end{align*}
This will stay true as long as $F(s, a_{\setminus i})$ is a function independent of $a_i$. Let $F(s, a_{\setminus i}) = - Q(s, a_{\setminus i}) - b(s, a_{\setminus i})]$ and combine it with the equation above:
\begin{multline*}
\nabla_{\theta_i} J (\theta) = \displaystyle \sum_{s \in D} d_\theta(s) \sum_{a \in \mathcal{A}} \Bigl[ \prod_{j \neq i} \pi_{\theta_j}(a_j | s) \Bigr] \nabla_{\theta_i} \pi_{\theta_i}(a_i | s) \cdot \\ [Q(s, a) - Q(s, a_{\setminus i}) - b(s, a_{\setminus i})]
\end{multline*}
\begin{multline*}
\quad \quad \quad = \displaystyle \sum_{s \in D} d_\theta(s) \sum_{a \in \mathcal{A}} \Bigl[ \prod_{j \neq i} \pi_{\theta_j}(a_j | s) \Bigr] \nabla_{\theta_i} \pi_{\theta_i}(a_i | s) \cdot \\ [Q_i(s, a_i) - b(s, a_{\setminus i})]
\end{multline*}
\begin{multline*}
\quad \quad \quad = \displaystyle \sum_{s \in D} d_\theta(s) \sum_{a \in \mathcal{A}} \pi_\theta(a|s) \nabla_{\theta_i} \log(\pi_{\theta_i}(a_i | s)) \cdot \\ [Q_i(s, a_i) - b(s, a_{\setminus i})].
\end{multline*}

Here, we prove that the policy gradient with respect to each $\theta_i$ can be obtained locally using the corresponding score function, $\nabla_{\theta_i} \log(\pi_{\theta_i}(a_i | s))$. By averaging $\theta_i^{(t)}$ from all agents updated by $Q_i(o_i, a_i) - b(o_i, a_{\setminus i})$, we obtain the same effect as updating the global $\theta$ based on $Q(s, a)$. Therefore, our method is as effective as IQL, but with a faster convergence rate due to parallel updates among all agents. Moreover, our framework for learning a homogeneous policy from local observation in the multi-agent learning framework is not restricted to MAPF and can potentially be applied to other MARL tasks, especially ones in partially observable settings.

\section{Experiments}
\label{sec:experiments}
In this section, we implement our methods\footnote{The code is available at https://github.com/Qiushi-Lin/SACHA.} and experimentally evaluate them with other methods on a server equipped with an Intel 2.3GHz 16-Core CPU and an NVIDIA A40 GPU. 

\subsection{Environment Setups}
\textbf{Training Environments:}
As mentioned above, our model is trained using the multi-agent actor-critic learning framework. Not only does each agent's policy network share parameters but also the critic networks applied to each subgroup of agents are homogeneous. We train our model over random grid maps of different sizes with randomly generated obstacles. The obstacle density is sampled from a triangular distribution between 0 and 0.5 with its peak value at 0.33. To fairly compare with other decentralized MAPF planners, our agent's policy network exclusively has $9 \times 9$ square-shaped FOV, the same as DHC and DCC, regardless of the environment size. Inspired by the curriculum learning~\cite{bengio2009curriculum}, we design a training pipeline that starts with only $2$ agents on $10 \times 10$ grid maps and gradually increases the number of agents and the size of the map once the success rate reaches a certain threshold. More and more complicated tasks are constantly added to the training pool until the map size exceeds $100 \times 100$ or the number of agents exceeds $72$.

\textbf{Testing Environments:}
We test our methods over a variety of maps all from the standard benchmark~\cite{stern2019multi}. We first select two random maps ($32 \times 32$ and $64 \times 64$) with uniformly distributed obstacles. Besides, we also use two game maps, $\textbf{den312d}$ ($65 \times 81$) and $\textbf{warehouse}$ ($161 \times 63$). The start-goal pairs of agent locations are randomly generated with the guaranteed existence of solutions. The number of agents is chosen from $\{4, 8, 16, 32, 64\}$ respectively. The maximum time step is $256$ for $\textbf{random32}$, $\textbf{random64}$, and $\textbf{den312d}$, and $512$ for $\textbf{warehouse}$. For cases that cannot be solved successfully within the time horizon or the runtime limit, we count each agent's step as the maximum time step.

\begin{figure}[t]
    \centering
    \includegraphics[width=0.77\columnwidth]{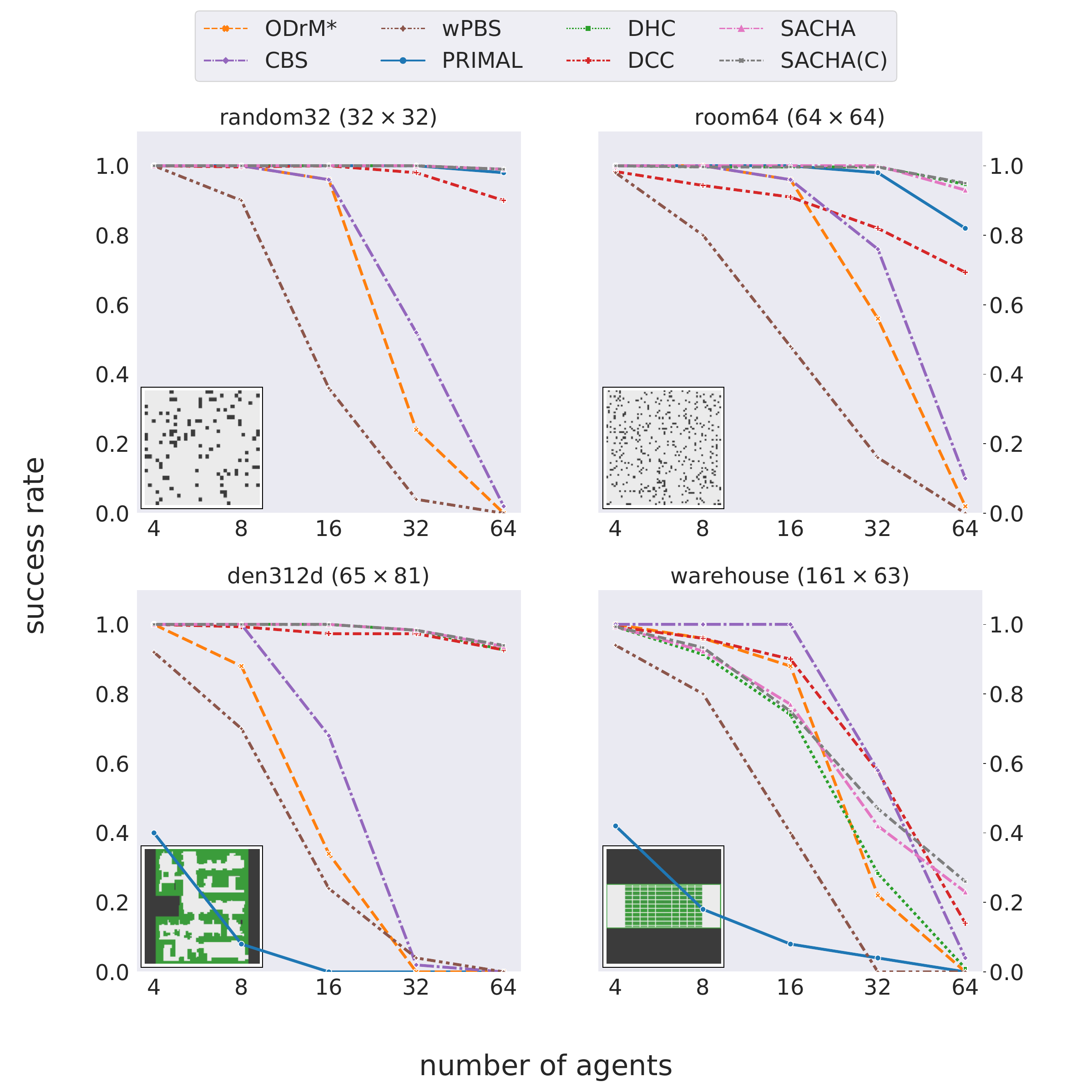}
    \caption{Success rates for different learning-based MAPF methods on different maps.}
    \label{fig:success_rate}
\end{figure}

\subsection{Baselines}
\textbf{Learning-Based Methods:}
We compare our methods with several state-of-the-art decentralized learning-based MAPF methods summarized in Table~\ref{tab:related_work}. PRIMAL uses expert demonstration from centralized MAPF planners to train its model. The expert demonstration has a positive effect on speeding up the training process but is very time-consuming and requires global information about the specific environment, which limits its generalization to unseen instances. DHC adopts IQL along with single-agent heuristic guidance and broadcast communication mechanism. The resulting model performed better than PRIMAL without any experts. DCC improves on DHC by learning selective communication with a decision causal unit, which can filter out redundant messages and focus on relevant information. This also reduces the communication frequency significantly.

\textbf{Search-Based Methods:}
Furthermore, we also compare our method with the centralized planners. We use two optimal but computationally expensive search-based methods for comparison: Conflict-Based Search (CBS)~\cite{sharon2015conflict} and ODrM$^*$~\cite{wagner2015subdimensional}. CBS performs a best-first search that expands the constraint tree node by adding constraints to each agent involved in every conflict, while ODrM$^*$ is a suboptimal planner based on M$^*$ that applies Operator Decomposition (OD) to split agents into independent conflict sets and thus reduce the complexity of joint planning. We also use Priority-Based Search (PBS)~\cite{ma2019searching} that searches for certain agent orders that can be used in the prioritized planning, which makes this solver incomplete but efficient. To simulate centralized planning in the same partially observable setting, we use windowed PBS (wPBS)~\cite{li2021lifelong}, which only avoids conflicts within a bounded time horizon. We set the time window length of wPBS to be equal to the caliber of the FOV to simulate the partially observable environments. We set the runtime limit of CBS and wPBS to $120$ seconds and ODrM$^*$ to $20$ seconds.

\subsection{Empirical Results}
We evaluate the performance of the MAPF methods based on two widely-used metrics, success rate and average step per agent. Success rate measures the ability to solve the given instances within the runtime limit, whereas the average step per agent measures the quality of the solutions over a given set of instances. We test our approach along with multiple baselines in around $300$ MAPF instances for each map with different numbers of agents.

Fig.~\ref{fig:success_rate} shows the success rate of our methods compared with all other baselines over four different MAPF maps. Within the time limit, all decentralized planners have a remarkable advantage in success rate. Even by including the precomputing time of shortest path heuristics, decentralized planners find solutions much faster than centralized ones. Among decentralized planners, PRIMAL tends to result in solutions with the worst quality, especially in those two game maps, which indicates that learning from the expert data cannot be easily generalized to instances with different numbers of agents and on maps with different structures. DHC and DCC have their advantages over PRIMAL by allowing the shortest path heuristics and the communication mechanism, although our methods can both outperform them in most cases. The advantages are more obvious over maps with higher obstacle density and more agents where more cooperation is demanded.

\begin{table}[t]
\centering
\caption{Solution quality for different MAPF methods.}
\label{tab:average_step}
\Huge
\resizebox{0.77\columnwidth}{!}{%
\begin{tabular}{c|c|cccccccc}
\hline
\multirow{2}{*}{Map} &
  \multirow{2}{*}{Agents} &
  \multicolumn{8}{c}{Average Step per Agent} \\ \cline{3-10} 
 &
   &
  \begin{tabular}[c]{@{}c@{}}CBS\\ (120s)\end{tabular} &
  \begin{tabular}[c]{@{}c@{}}ODrM*\\ (20s)\end{tabular} &
  \multicolumn{1}{c|}{\begin{tabular}[c]{@{}c@{}}wPBS\\ (120s)\end{tabular}} &
  \begin{tabular}[c]{@{}c@{}}PRI\\ MAL\end{tabular} &
  DHC &
  DCC &
  \begin{tabular}[c]{@{}c@{}}SAC\\ HA\end{tabular} &
  \begin{tabular}[c]{@{}c@{}}SAC\\ HA(C)\end{tabular} \\ \hline
\multirow{5}{*}{\rotatebox[origin=c]{90}{\textbf{random32}}} &
  4 &
  21.82 &
  21.82 &
  \multicolumn{1}{c|}{22.90} &
  32.96 &
  35.70 &
  32.83 &
  \textbf{29.93} &
  31.03 \\
 &
  8 &
  21.38 &
  21.37 &
  \multicolumn{1}{c|}{46.06} &
  38.62 &
  42.64 &
  39.56 &
  \textbf{36.34} &
  38.30 \\
 &
  16 &
  31.16 &
  31.26 &
  \multicolumn{1}{c|}{172.12} &
  45.12 &
  48.67 &
  43.56 &
  41.71 &
  \textbf{41.30} \\
 &
  32 &
  133.86 &
  199.47 &
  \multicolumn{1}{c|}{246.61} &
  50.34 &
  52.17 &
  56.11 &
  50.26 &
  \textbf{47.72} \\
 &
  64 &
  251.30 &
  \sout{256.00} &
  \multicolumn{1}{c|}{\sout{256.00}} &
  69.40 &
  \textbf{66.05} &
  88.79 &
  76.47 &
  74.48 \\ \hline
\multirow{5}{*}{\rotatebox[origin=c]{90}{\textbf{random64}}} &
  4 &
  42.94 &
  42.95 &
  \multicolumn{1}{c|}{48.14} &
  67.82 &
  71.04 &
  70.80 &
  \textbf{65.47} &
  67.10 \\
 &
  8 &
  42.74 &
  42.80 &
  \multicolumn{1}{c|}{84.52} &
  74.68 &
  82.43 &
  88.94 &
  \textbf{70.49} &
  72.38 \\
 &
  16 &
  51.51 &
  51.52 &
  \multicolumn{1}{c|}{154.47} &
  89.22 &
  94.22 &
  102.27 &
  83.74 &
  \textbf{82.17} \\
 &
  32 &
  94.36 &
  136.67 &
  \multicolumn{1}{c|}{222.08} &
  98.02 &
  103.05 &
  126.71 &
  95.67 &
  \textbf{93.08} \\
 &
  64 &
  234.66 &
  251.65 &
  \multicolumn{1}{c|}{\sout{256.00}} &
  105.12 &
  120.68 &
  154.72 &
  99.02 &
  \textbf{96.42} \\ \hline
\multirow{5}{*}{\rotatebox[origin=c]{90}{\textbf{den312d}}} &
  4 &
  51.74 &
  51.76 &
  \multicolumn{1}{c|}{69.32} &
  196.54 &
  86.56 &
  82.99 &
  \textbf{78.33} &
  81.43 \\
 &
  8 &
  55.50 &
  78.74 &
  \multicolumn{1}{c|}{116.32} &
  245.02 &
  100.70 &
  97.95 &
  \textbf{84.24} &
  89.73 \\
 &
  16 &
  118.97 &
  186.44 &
  \multicolumn{1}{c|}{208.28} &
  \sout{256.00} &
  109.24 &
  108.29 &
  97.86 &
  \textbf{96.74} \\
 &
  32 &
  251.86 &
  \sout{256.00} &
  \multicolumn{1}{c|}{248.06} &
  \sout{256.00} &
  124.38 &
  119.15 &
  111.28 &
  \textbf{104.30} \\
 &
  64 &
  \sout{256.00} &
  \sout{256.00} &
  \multicolumn{1}{c|}{\sout{256.00}} &
  \sout{256.00} &
  153.17 &
  145.21 &
  \textbf{140.79} &
  142.97 \\ \hline
\multirow{5}{*}{\rotatebox[origin=c]{90}{\textbf{warehouse}}} &
  4 &
  77.79 &
  77.79 &
  \multicolumn{1}{c|}{104.41} &
  355.80 &
  146.12 &
  135.89 &
  \textbf{131.43} &
  134.59 \\
 &
  8 &
  83.48 &
  100.37 &
  \multicolumn{1}{c|}{170.46} &
  451.82 &
  198.82 &
  169.50 &
  \textbf{164.83} &
  166.72 \\
 &
  16 &
  81.64 &
  133.59 &
  \multicolumn{1}{c|}{340.18} &
  492.04 &
  281.37 &
  208.72 &
  \textbf{192.30} &
  198.72 \\
 &
  32 &
  262.15 &
  417.22 &
  \multicolumn{1}{c|}{\sout{512.00}} &
  505.58 &
  432.28 &
  \textbf{335.81} &
  370.65 &
  354.33 \\
 &
  64 &
  494.93 &
  \sout{512.00} &
  \multicolumn{1}{c|}{\sout{512.00}} &
  \sout{512.00} &
  \sout{512.00} &
  473.92 &
  449.83 &
  \textbf{437.29} \\ \hline
\end{tabular}%
}
\end{table}

Table~\ref{tab:average_step} reports the average step required to finish goals from each agent in multiple different instances. If planners exceed the runtime limit in some cases, we consider them failures and count them as spending the maximum time horizon. The stroke-out data with maximum time steps indicates zero success. When compared to search-based planners with relatively small numbers of agents, as expected, all learning-based methods cannot provide comparable results. However, as the number of agents grows, the search-based methods are significantly more time-consuming than learning-based methods and the success rate would drop rapidly due to the runtime limit. The two communication-based methods DHC and DCC have greater solution quality over PRIMAL in all cases. However, SACHA and SACHA(C) outperform them by a decent margin in most instances, with and without the communication block, which demonstrates their advantages over other learning-based methods. It is worth mentioning that, generally, SACHA(C) has better performance than SACHA in instances with larger numbers of agents where communication can be rather helpful.

As reported in \cite{ma2021distributed}, DHC always has better performance than its alternative without the communication unit. Hence, it shows that our method can serve the non-communicating scenarios when SACHA can outperform DHC and thus DHC without communication. Besides, DHC can essentially be viewed as training our communication-based model without the attention block via the independent $Q$-learning. Therefore, the fact that SACHA(C) has greater performance than DHC demonstrates the strength of the heuristic-based attention mechanism and the multi-agent learning framework. Overall, our methods have a better chance to solve given MAPF instances, with and without communication, and among those solved instances, they result in solutions with better quality.

\section{Conclusion}
In this paper, we introduced SACHA, a novel approach for learning cooperative policies for MAPF and potentially other MARL problems in partially observable environments. SACHA combines the multi-agent soft actor-critic that maximizes both expected reward and entropy with heuristic-based attention mechanisms that enhance the network architectures of both the actor and the critic. Specifically, we proposed to augment each agent's local observation with heuristic guidance from other agents and to use an attention module that learns to focus on the most relevant information for each agent to avoid collisions and achieve the goal. To the best of our knowledge, we are the first to apply the soft actor-critic with a novel agent-centered critic to homogeneous MARL settings with partial observability and to incorporate heuristic guidance and attention in the agent's policy network. We evaluated our method on various MAPF benchmarks and showed that it outperforms existing baselines for almost all the cases in terms of success rate and solution quality.


\bibliographystyle{IEEEtran}
\bibliography{references}

\end{document}